# CMSG: Cross-Media Semantic-Graph Feature Matching Algorithm for Autonomous Vehicle Relocalization

Shuhang Tan
Hefei Institutes of Physical Science, CAS, 230031
University of Science and Technology of China, 230026
Hefei, China,
stan9177@mail.ustc.edu.cn

Hengyu Liu
Dept. of Information and Communication
Xiamen University
Xiamen, China
23320201154007@stu.xmu.edu.cn

Zhiling Wang*
Hefei Institutes of Physical Science, 230031
Chinese Academy of Sciences
Hefei, China
zlwang@hfcas.ac.cn

**Abstract**. Relocalization is the basis of map-based localization algorithms. Camera and LiDAR map-based methods are pervasive since their robustness under different scenarios. Generally, mapping and localization using same sensor has better accuracy since matching feature between same type of data is easier. However, due to camera's lack of 3D information and the high cost of LiDAR, cross-media methods are developing, which combined live image data and Lidar map. Although matching features between different medias is challenging, we believe cross-media is the tendency for AV relocalization since its low-cost and the accuracy can be comparable to the same-sensor based methods. In this paper, we propose CMSG, a novel cross-media algorithm for AV relocalization task. Semantic features are utilized for better interpretating the correlation between point clouds and image features. What's more, abstracted semantic graph nodes are introduced, and a graph network architecture is integrated to better extract the similarity of semantic feature. Validation experiments are conducted on the KITTI odometry dataset. Our results show that CMSG can have comparable or even better accuracy compared to current single-sensor based methods at a speed of 25 FPS on NVIDIA 1080 Ti GPU. Related code can be found in https://github.com/Clouds1997/CMSG.

## 1. Introduction

Accurate localization is the key part of the real autonomous vehicle (AV) system, ego cars depend on it to achieve their current positions, which is crucial for detection and decision system performance. A typical localization pipeline contains two parts [1]. First, a rough pose of the ego car is estimated, giving observers a view of global localization. The second part will refine the previous rough results and provide more accurate local localization. The first part will provide a reference point for selecting the landmarks that will contribute to the subsequent finer localization process [2], which can reduce the computation load and accelerate the whole localization process. This is also called localization initialization or relocalization and is the basis of map-based localization methods. GPS/GNSS-based methods are usually used for the first part since they are low-cost, efficient positioning solutions [3]. However, their errors can be of the order of tens of meters [4] in environments with low signals, which are unacceptable for real AV applications.

Vision-based [5-8] and LiDAR (light detection and ranging)-based [9-12] methods have been proposed to overcome the GPS limitation since they are more reliable and can provide higher accuracy under various scenarios. These two types of methods usually depend on high-definition (HD) maps, which are built with LiDAR or vision data in advance. Real-time scanning (laser or visual) features are matched to the pre-built HD map to provide an accurate pose. Vision-based methods can provide a more reasonable accuracy and the camera is economic. However, building a 3d map with a camera is hard due to its restricted range of view and the lack of 3D information. LiDAR-based methods are more accurate and robust. But the cost of LiDAR is consuming, so it is impossible to mass install them on cars. Meanwhile, its computation load is heavy. Cross-media methods arise [13,14] due to the shortage of single-sensor methods, which can make use of the advantages of both LiDAR and camera. Most of them use an onboard camera to collect real-time vision data, then match the vision features with a pre-built 3D LiDAR map. Unfortunately, matching features across the LiDAR and camera is not a handy problem, and this challenge causes the performance of cross-media methods inferior to single-sensor methods.

Considering the specific 3D space information LiDAR can provide and the low cost of a camera, cross-media methods have an undeniable advantage. However, due to the difficulty of addressing the cross-media feature matching problem, there is no perfect method has been proposed now. How to match vision features in the LiDAR map efficiently and output a reasonable global localization result is still an open question. In this paper, we proposed a new learning-based Cross-Media Semantic-Graph (CMSG) relocalization algorithm to tackle the difficulty of image and point cloud feature matching. To be specific, AVs only need onboard cameras to get real-time environment data when the LiDAR map was pre-built and the feature of point clouds can be extracted offline, which can effectively lower the system cost. Semantic features of the image are extracted in real-time and fused with pre-extracted semantic point cloud features, which can provide finer information and improve the correlation between two modules. A graph network is introduced to calculate the relationship of semantic graph nodes, and this further improves the robustness and accuracy of the algorithm. To our best knowledge, we are the first ones who apply the semantic-graph representation and graph network to the cross-media relocalization task. Our algorithm has been validated on the KITTI odometry dataset (a project of Karlsruhe Institute of Technology and Toyota Technological Institute at Chicago) using F1 score. The accuracy of CMSG can be comparable to current single sensor-based algorithms with an obvious economic advantage, which proved its effectiveness and reliability.

The rest of the paper is organized as follows: Section 2 simply analyzes the recent related map-based feature matching methods. Section 3 describes the specific CMSG model pipeline. Section 4 provides and analyzes the results of the experiments using the KITTI dataset. Section 5 gives the conclusion of the whole paper and discusses future work.

2. **Related Works**

Mapping and collection live data using only camera or LiDAR is pervasive because they are the most accessible sensors for current AV system. Generally, they can provide better results than using multi-type sensors since feature matching from same sensor is more easily. Du et al. [5] proposed an effective lane lines extraction method with vision-based map, which can provide precise information for further localization process. Yoneda et al. [6] proposed the localization method based on a mono-camera and a digital map, which used image temple matching between map and live image data. Li et al. [8] proposed a new image-level localization method with one single monocular camera, including a holistic feature matching using vision-based map. Wen et al. [11] proposed a localization method based on LiDAR lightweight pre-built occupancy HD map. Charroud et al. [12] proposed a deep learning based framework to predict the position with a few point clouds. Chiang et al. [15] proposed new strategies to cope with the localization with LiDAR and an improved NDT pose estimation method. Feature matching and localization method based on same sensor are reliable and effective. However, camera cannot provide valuable spatial and 3D information, which makes it is difficulty to build precise 3D map. The cost of LiDAR is still too high to mass install it on all AV system, and the point clouds data need large computing source. Theses disadvantages hinder the development of localization methods using same sensor.

To better exploit the low cost of camera and the abundant 3D information of LiDAR data, cross-media localization methods have appeared. Usually, a sensor-rich AV system is used to build LiDAR map offline. Production AV systems only need light-weight camera sensors, and the live image data is used to match the feature in pre-built LiDAR map, performing a rough global localization. Kim et al. [16] utilized the depth information of a stereo camera and matched the camera feature with offline LiDAR map to realize localization when GPS denied. Yoneda et al. [14] proposed a localization method using image temple matching between image data and predefined LiDAR map. Yu et al. [17] combined geometric 3D lines in LiDAR map and 2D line in video to mitigate the feature gap and to realize the localization based on a monocular camera. Previous works only use the raw data to execute the feature matching. With the development of computing system and semantic technology, semantic information has been considered into in some works with the hypothesis that semantic can provide more invariable appearance information. Yin et al. [13] proposes a localization method using LiDAR and BIM-generated maps considering both geometric and semantic properties. Qin et al. [18] applied visual semantic map and the map data can be updated by on-cloud road data, which can assist user-end cars to self-localize. Wang et al. [19] proposed a relocalization method using local semantic descriptor, which combined normal and spatial relationships.

Although many efforts have been made about semantic information and multi-media feature matching, they do not explore it thoroughly. In this paper, we focus on relocalization task and proposed a cross-media network using both semantic LiDAR map and image data. A graph network is integrated to better extract the feature nodes relationship in semantic image. LiDAR and image feature are combined in the end to infer the result, giving a global localization for ego-car.

### 3. Proposed CMSG Design Method

3.1. CMSG Pipeline Overview

The whole CMSG relocalization network can be divided into LiDAR map and RGB image two process parts. Figure 1 shows the whole framework of the CSGM network. Point clouds data is processed offline in advance and the feature of the semantic LiDAR map is extracted by two separate branches. One branch encodes semantic point feature, and the other uses a graph network to extract the semantic image node relationship. Image data is collected and processed in real-time, there are also two input branches. Raw image data and semantic image are passed by image encoder and same graph network is used for semantic image. Image and LiDAR features are fused at different stage of the network for results inference.

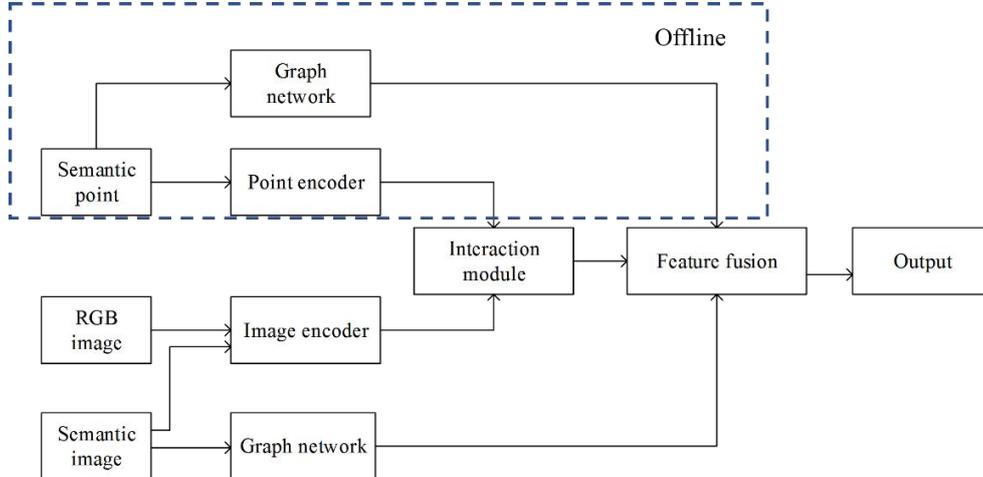

**Figure 1.** The whole framework for CSGM.

The following of the section will go through each part of the CSGM network in Figure 2. (A) The point clouds and image feature extraction and fusion module, which contains point encoder, image encoder and attention feature fusion, drawn by green boxes in Figure 2. (B) The Graph network which is used to extract the semantic node relationship of LiDAR and image separately, drawn by yellow boxes in Figure 2. (C) All extracted features are processed by a shared multilayer perceptron (MLP) layer and go through fully connected (FC) layers to give final classification results, shown in gray boxes in Figure 2.

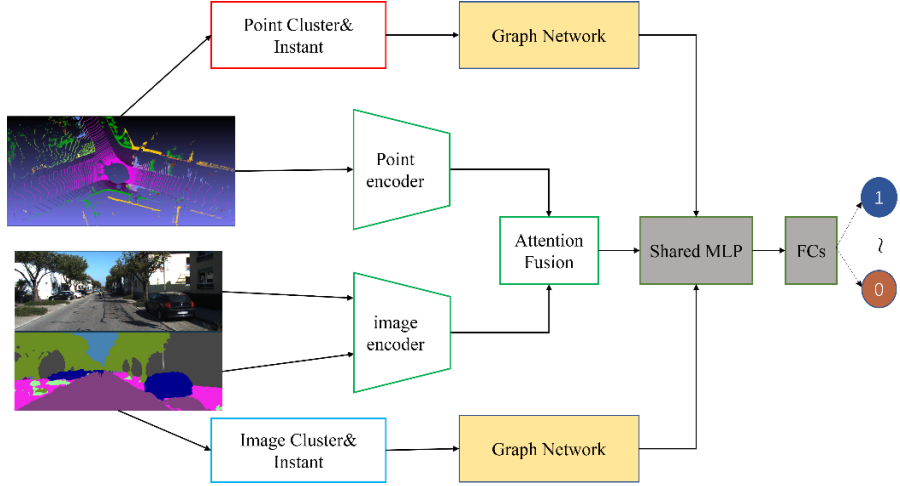

**Figure 2.** Detailed network description of CSGM.

### 3.2. Multi-Media Feature Extraction and Fusion Module

LiDAR can provide precise 3D spatial information while image can provide finer colour information in 2D planer. However, there is no strong correlation between spatial and colour information because the mapping relationship between point clouds and image is not clear. Thus, an effective method needs to be found to match the features between point clouds and image. Another challenge is that the inconsistent depth information in point clouds and image, which makes it hard to match the features between them directly.

During the cross-media feature matching process, different types of the information will influence each other. Inspired by Deepi2p [20], we adopt the same encoder network to extract and fuse the point clouds and image features. The encoder network includes three parts, point clouds encoder, image encoder and feature fusion module.

(a) Point clouds encoder: This part combines So-net [21] and Pointnet++ [22]. In our work, besides the spatial information, the semantic information of point clouds is also introduced to provide more invariant instance information.

(b) Image encoder: Same as Resnet-34 [23], and it also combines the raw image and semantic image information.

(c) Feature fusion module: This is an attention fusion module and its input $(F_p, F_i, F_h)$ contains point clouds' global feature $F_p$, image's global feature $F_i$ and high dimensional feature of image $F_h$. The attention score can be obtained by (1).

$$Satt = f(F_p, F_i) \qquad (1)$$

Where $f(g)$ expresses the fusion function, it can be addition, multiplication and other mathematical functions.

$$F_{wi} = g(Satt, F_h) \qquad (2)$$

(2) shows the function to get the weighted image feature $F_{wi}$ which integrates $Satt$ and $F_h$. $g(g)$ is the weight function chosen by researchers. Finally, the interaction between two media features can be realized through $F_{wi}$ and $F_p$.

### 3.3. Graph Network

#### 3.3.1. LiDAR Semantic Node Extraction

The instances in semantic LiDAR data are extracted by clustering, and each instance is considered as a semantic image node, which expresses the scene information of point clouds. To better extract the instance node, we adopt a distance adaptive clustering algorithm. This algorithm considers the spatial distribution characteristics of LiDAR data.

Generally, point clouds are denser when close to LiDAR, so smaller clustering radius is used to cluster the instance, while larger radius is used for clustering further point clouds. Supposed that there are $N$ points need to be clustered. For each point $p_i$, we calculate the distance $dist(p_i, p_j)$ between it to its nearest neighbour point $p_j$. The clustering radius $R(p_i)$ is generated by (3).

$$R(p_i) = \alpha * dist(p_i, p_j) \quad (3)$$

Where $\alpha$ is a coefficient to adjust the result. Then, if $dist(p_i, p_j) <= R(p_i)$, these points are be clustered into the same instance.

### 3.3.2. Image Semantic Node Extraction

For image semantic node extraction, we choose to obtain the semantic label form the semantic image directly, which can guarantee the instantaneity. Similar to [24,25], the semantic information of the semantic image is clustered to get the instances. The instance can be expressed by its central coordinate $(u_i, v_i)$. Each node contains the coordinates and semantic label information.

### 3.3.3. Graph Similarity Network

For better applying CMSG to AV relocalization task, the computation efficient is of great concern. Meanwhile, multi-view variation is common during AV running. However, the sort of the calculated similarity scores should not change. Based on above requirements, we refer to Simgnn [26], graph similarity network is introduced into cross-media feature matching task. Considering the effectiveness [27] of DGCNN [28], we adopt its EdgeConv structure for the semantic image node feature encoding to get the local feature. The position information of the node is used by K-nearest neighbour clustering algorithm, then the feature of each cluster is calculated. To be specific, $l_i$ is the semantic feature of one specific node, the k-nearest node around it can be defined $l_{mi}$, where $m \in \{1, 2, ...k\}$. The aggregation feature in this set can be expressed as (4).

$$h_\Theta(f_i, f_j) = \overline{h}_\Theta(f_i, f_i - f_j^m) \quad (4)$$

Where $\Theta$ is the learning coefficient for aggregation feature extraction. Then the local relationship of centre point feature $l_i$ and neighbourhood feature $K$ can be learned.

Inspired by [26,27], a learnable matrix is introduced to screen valuable node. For example, a semantic image with $N$ nodes, each node is defined as $n_i$. Specifically, the global feature $g$ is calculated by averaging the feature of all nodes, shown in (5).

$$g = \frac{1}{N} \sum_{i=1}^{N} n_i \cdot M \quad (5)$$

$N$ is the number of the nodes; $n_i \in R^D$ represents the feature of node $i$; $M \in R^{D \times D}$ is the learnable matrix, which can be used to dynamic adjust each node's effect on global feature. The attention score $Att_{score}$ can be calculated by (6) using hyperbolic tangent function $\tanh(g)$ and *Sigmoid* function.

$$Att_{score} = sigmoid(n_i \cdot \tanh(g)) \quad (6)$$

The final feature $F_{whole}$ of the whole semantic image can be obtained by multiplying attention score $Att_{score}$ and each node's feature $n_i$, shown in (7).

$$F_{whole} = \sum_{i=1}^{N} Att_{score} \cdot n_i \quad (7)$$

### 3.3.4. Feature Fusion Module

Multi-modal features got above are fused by a shared multi-layer perception layer, followed by a couple of fully connected layers to down-sample the dimension of feature vectors. The task can be simplified to a binary classification according to the similarity result $y \subseteq [0,1]$. The loss function is optimized by classic binary cross-entropy loss function (8).

$$L = -\sum_{i=1}^{N} p_i \log q_i \qquad (8)$$

## 4. Experiments Results and Analysis

### 4.1. Dataset and Implementation Details

We use the KITTI [29] odometry dataset with F1 score as the evaluation, which has been widely used for relocaliation and cross-media matching tasks [20, 27]. KITTI odometry contains total 22 stereo sequences, 11 sequences of them (00-10) have ground truth and the rest of them without ground truth. It provides image and LiDAR data with label which can meet our cross-media matching experiment requirements. Also, combining with KITTI semantic dataset, fine semantic information can be obtained and integrated easily, which the basis of CMSG. What's more, KITTI odometry is also suitable for our future localization work, the next step of the relocalization, which can make our work consistent.

In this paper, we select sequences with ground truth to evaluate the proposed relocalization algorithm, the ground truth is used to verify if the point clouds and image matching successful. And a pair of positive samples is established when the Euclidean distance between image and point clouds is smaller than 2 meters. While the negative sample pair is when the Euclidean distance bigger than 20 meters for clarification, considering we cannot decide if the feature pair belongs to one same scene when their Euclidean distance is between 2 meters and 20 meters. The sequences with repeated road lines (00, 02, 05, 06, 07, 08) are used, which can provide more samples.

Rangenet++ [30] and deeplabv3+ [31] are used for point clouds and image semantic segmentation. Figure 3 shows the illustration of semantic segmentation results for image and point clouds, more instances are integrated to ensure the consistent of two modules' semantic information. The number of extracted semantic graph nodes is set to constant for different scenes, which is set to 35 for experiment in this paper. For graph with insufficient nodes, virtual nodes are inserted, and for graph with exceeding nodes, required nodes are selected by random sampling. Similar to [27], we select one sequence as the test set and the rest as the training set. Then exchanging the test set with the training set until all the sequences have been tested. The whole network of our CSGM is end-to-end optimized by Adam optimizer. For all the training set, the network is trained for 35 epochs on 4 K80 GPUs, with learning rate starts at 0.001.

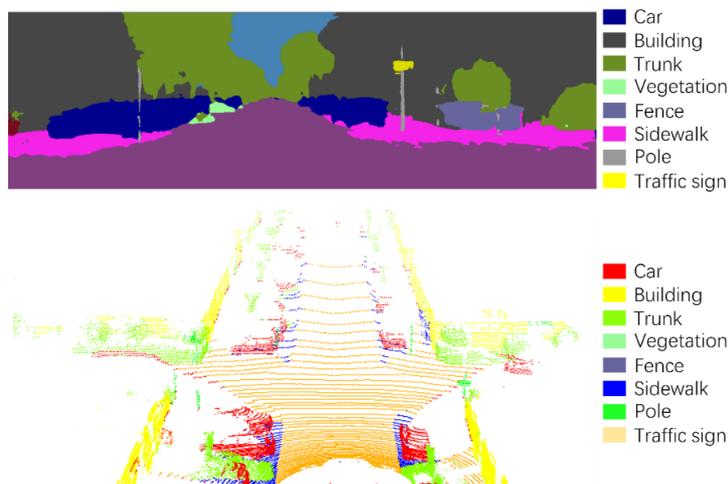

**Figure 3.** Illustration of semantic segmentation result.
The upper one belongs to image and the lower one belongs to point clouds.

4.2. Results for Cross-Media Feature Matching

Because there is little open-sourced cross-media matching algorithms, and there is no existing leaderboard for relocalization task. So we compare our algorithm performance with latest open-sourced single LiDAR sensor algorithms which are also evaluated using KITTI odometry and F1 score (M2DP2 [32], Scan Context3 (SC) [33] and PointNetVLAD [34]). The algorithm performance is quantitative analysed and compared by F1 score (9).

$$F_1 = 2 \times \frac{P \times R}{P + R} \quad (9)$$

Where $P$ is the precision, $R$ represents the recall rate. The final results are shown in Table 1.

**Table 1.** The F1 results of selected algorithms.

| Algorithm | 00 | 02 | 05 | 06 | 07 | 08 | Mean |
|---|---|---|---|---|---|---|---|
| [32] | **0.836** | **0.781** | 0.772 | 0.896 | 0.861 | 0.169 | 0.719 |
| [33] | *0.937* | *0.858* | *0.955* | *0.998* | *0.922* | *0.811* | *0.914* |
| [34] | 0.785 | 0.710 | 0.775 | **0.903** | 0.448 | 0.142 | 0.627 |
| Ours (CMSG) | **0.836** | 0.774 | **0.825** | 0.770 | **0.863** | *0.826* | **0.816** |

All algorithms are tested on 00, 02, 05, 06, 07, 08 sequences, and the highest F1 score is labelled by red, and the second highest score is labelled by blue. As Table 1 shown, our algorithm demonstrates most the second highest F1 score on KITTI dataset, which lowers than Scan Context3 (SC) [33] performance but higher than other two single-LiDAR [32,34] based methods.

It is worth to be notified that since the LiDAR provides 360° panorama information while image provides a narrower view, images input much less valuable information than point clouds. Therefore, our cross-media algorithm shows an inferior performance than single-LiDAR methods on many sequences. However, except SC [33], CMSG shows much better and robustness than [32, 34]. And CMSG even get top-one at sequence 08, 08 has more similar scene graphs than other sequences, which causes the failure of single-LiDAR based methods. And this proves our cross-media algorithm's rationality and superiority. What's more, our CMSG displays a stable performance on all the sequences, without apparent fluctuation exiting in [32, 34]. We believe this is benefit by the semantic information, which is also used by SC [33]. And its stable performance also verified the effectiveness of semantic information.

4.3. Ablation Study

Table 2 shows the influence of all proposed modules for CMSG's effectiveness and accuracy. Each architecture trained and tested on sequence 00 and the results were evaluated and compared using F1 scores.

**Table 2.** Performance of proposed method with different modules,

w means include this module, w/o means without.

| Methods | Base | Semantic | Graph network | F1 score |
|---|---|---|---|---|
| (a) | w | w/o | w/o | 0.759 |
| (b) | w | w | w/o | 0.782 |
| (c) | w | w | w | 0.836 |

Method (a) is the base architecture without semantic information and graph network, and its F1 score is 0.759. Method (b) adds semantic information as the input, its F1 score increases to 0.782. Method (c) integrates both semantic input and graph network and the F1 score reaches 0.836, which shows an obvious improvement.

Semantic information provides specific correlation between different medias. For example, "tree" in images also has "tree" label in point clouds, which is helpful for cross-media feature matching. However, different semantic algorithms for image and point clouds also cause noise information. For this point, redundant information can be filtered by abstracted semantic graph nodes. And graph network can extract more higher-level features.

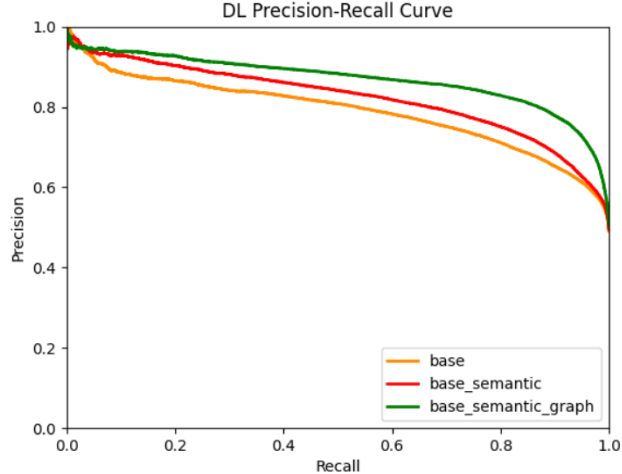

**Figure 4.** P-R curve for method (a) (b) (c).

Figure 4 shows the Precision-Recall (P-R) curve for all the methods mentioned in Table 2, and the curve visually shows the performance improvement when base architecture adds semantic information and graph network. The results prove the reasonability and effectiveness of semantic and graph modules we proposed in the paper.

4.4. Algorithm Efficiency Test

Considering our algorithm is for real autonomous vehicles task, which asks for proper model size and efficiency. But relocalization is different from real-time localization task, this task is mainly for system initialization, which can tolerate a little delay. The inference time of CMSG on single 1080Ti can be within 40ms and the model size is about 100MB, which can meet the real-time and model deployment requirement at some real AV hardware systems.

## 5. Conclusion

In this paper, we present CMSG, a novel cross-media-based network with live image input and a pre-built LiDAR map for AV relocalization tasks. The LiDAR map is established and its semantic feature is extracted in advance for efficient and economic reasons, image scene information is collected in real-time. Semantic point clouds and image features are used as input to provide more invariant instance features CSGM first encodes the two medias' semantic feature and the raw RGB image, then these features are fused to learn more valuable information. A graph network is utilized to calculate the structural similarity of the semantic graph, and all the features are fused at the end part of the network. The final fused feature map goes through a shared MLP and a couple of fully connected layers to classify the results. By taking full advantage of LiDAR and image representations, our CMSG strikes a careful balance of accuracy, efficiency, and economy. The encouraging results on the KITTI odometry dataset demonstrate out CMSG can guarantee a stable and reasonable performance on AV relocalization tasks. And to further facilitate real-time localization tasks and downstream tasks.

**ACKNOWLEDGMENTS**
This work was supported in part by Key Science and Technology Project of Anhui (Grant No. 202103a05020007) and National Key Research and Development Program of China (No. 2020AAA0108103).

for Visual Localization towards Autonomous Driving. In 2021 IEEE International Conference on Robotics and Automation (ICRA) (pp. 11248-11254). IEEE.